\documentclass{article}
\usepackage{spconf,amsmath,graphicx}
\usepackage{color}
\usepackage{subfigure}

\usepackage{multirow}

\def\ie{\textit{i.e.}}

\def\eg{\textit{e.g.}}

\title{MISA: Unveiling the Vulnerabilities in Split Federated Learning}
\name{
\parbox{\textwidth}{\centering{Wei Wan$^{1,5,6,7,8}$, Yuxuan Ning$^{2}$, Shengshan Hu$^{1,5,6,7,8}$, Lulu Xue$^{1,5,6,7,8}$,\\
Minghui Li$^{3}$, Leo Yu Zhang$^{4}$, and Hai Jin$^{2,5,6,9}$}}  \vspace{-0.2cm}}

\address{
$^1$ School of Cyber Science and Engineering, Huazhong University of Science and Technology\\
$^2$ School of Computer Science and Technology, Huazhong University of Science and Technology\\
$^3$School of Software Engineering, Huazhong University of Science and Technology\\
$^4$ School of Information and Communication Technology, Griffith University\\
$^5$ National Engineering Research Center for Big Data Technology and System\\
$^6$ Services Computing Technology and System Lab \\
$^7$ Hubei Key Laboratory of Distributed System Security\\
$^8$ Hubei Engineering Research Center on Big Data Security $^9$ Cluster and Grid Computing Lab
}
\begin{document}
%
\maketitle
\begin{abstract}
\textit{Federated learning} (FL) and \textit{split learning} (SL) are prevailing distributed paradigms in recent years. They both enable shared global model training while keeping data localized on users' devices. The former excels in parallel execution capabilities, while the latter enjoys low dependence on edge computing resources and strong privacy protection. \textit{Split federated learning} (SFL) combines the strengths of both FL and SL, making it one of the most popular distributed architectures. Furthermore, a recent study has claimed that SFL exhibits robustness against poisoning attacks, with a fivefold improvement compared to FL in terms of robustness.

In this paper, we present a novel poisoning attack known as MISA. It poisons both the top and bottom models, causing a \textbf{\underline{misa}}lignment in the global model, ultimately leading to a drastic accuracy collapse. This attack unveils the vulnerabilities in SFL, challenging the conventional belief that SFL is robust against poisoning attacks. Extensive experiments demonstrate that our proposed MISA poses a significant threat to the availability of SFL, underscoring the imperative for academia and industry to accord this matter due attention.
\end{abstract}
\begin{figure*}[t]
\centerline{\includegraphics[width=1.96\columnwidth]{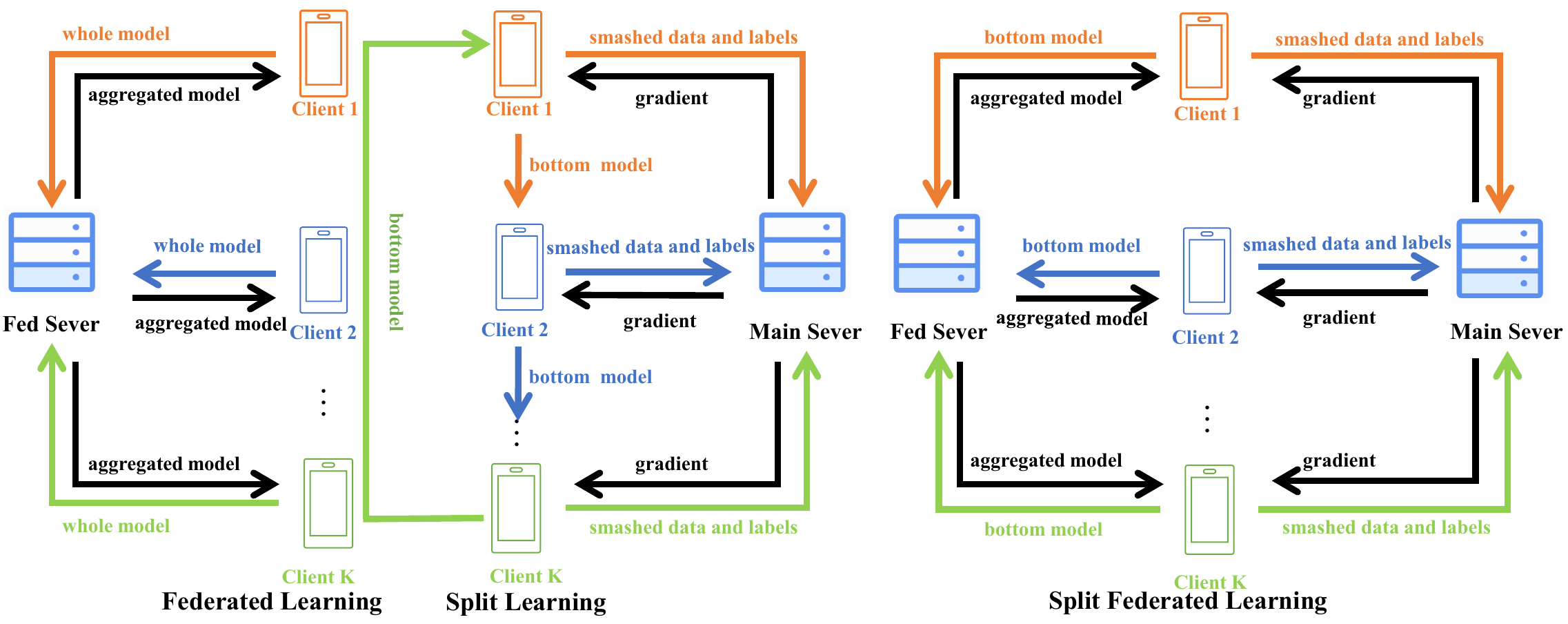}}
\vspace{-3mm}
\caption{The overview of FL, SL, and SFL}
\vspace{-5mm}
\label{fig:SFL}
\end{figure*}

\vspace{-4mm}
\section{Introduction}
\label{sec:intro}
\textit{Federated learning} (FL)~\cite{FedAvg,FLSurvey} and \textit{split learning} (SL)~\cite{SL} have emerged as dominant distributed paradigms in recent years. They enable multiple participants to collaboratively build a shared global model under the coordination of a central server, without directly uploading training data to a third party. However, FL poses challenges for edge clients with limited computational resources since it requires them to complete the full model training. Additionally, FL necessitates the upload of complete model parameters by clients, making it easy for attackers to infer the training data of clients, leading to privacy concerns~\cite{FLDPGAN}. On the other hand, SL permits only one client to communicate with the central server per round, preventing multi-client parallel training and thereby resulting in lower efficiency.

To address these challenges, \textit{split federated learning} (SFL)~\cite{SplitFed} has been proposed. This paradigm combines FL and SL, effectively resolving issues related to limited computational resources at edge clients and privacy concerns. Moreover, it permits parallel execution by multiple clients, significantly enhancing efficiency. Fig.~\ref{fig:SFL} offers an overview of FL, SL, and SFL.

In addition to the aforementioned advantages, SFL is also shown to exhibit strong resilience against poisoning attacks~\cite{SFLAGRT,SFLDataPoisoning,SFLDataPoisoning2}. This resilience stems from the model's division into two parts, with clients having control over only one portion of the model parameters (\ie, bottom model), significantly reducing the attacker's exploitable space. Based on this insight, existing research has largely overlooked the exploration of the robustness of SFL. To the best of our knowledge, related work has been limited to \textbf{\underline{t}}ailoring bottom model poisoning attacks according to the \textbf{\underline{ag}}g\textbf{\underline{r}}egation algorithm (AGRT)~~\cite{SFLAGRT} and uploading false labels~~\cite{SFLDataPoisoning,SFLDataPoisoning2}.

To garner attention from both academia and industry regarding research on the robustness of SFL, we propose an advanced poisoning attack known as MISA. It poisons both the top and bottom models, causing a misalignment in the global model, ultimately leading to a drastic accuracy collapse. For the top model, we meticulously design perturbations for smashed data, carefully balancing attack stealthiness and effectiveness, leading to indirect poisoning. Regarding the bottom model, we formalize its poisoning as a Min-Sum optimization. To solve this optimization, we introduce Thompson Sampling~\cite{thompsonsampling} for dynamically determining the optimal perturbation direction. Additionally, we propose SnL (\textit{search and locate}), which combines coarse-grained searching and fine-grained locating to obtain the optimal perturbation magnitude.

The key contributions of this paper can be encapsulated as follows:

\vspace{-2mm}
\begin{itemize}
    \item  We introduce a highly potent poisoning attack called MISA, challenging the conventional belief that SFL is robust against poisoning attacks. Concurrently, we call upon the academic and industrial communities to pay significant attention to the vulnerability of SFL.

   \vspace{-2mm}
    \item To the best of our knowledge, MISA is the first method to simultaneously poison both the top model and the bottom model, inducing a misalignment in the global model, and ultimately leading to a collapse in accuracy.

\vspace{-2mm}
    \item We conduct extensive experiments to evaluate MISA. The results indicate that, compared to SOTA attacks, MISA can achieve a higher accuracy drop. Furthermore, we also investigate the impact of varying model split positions, attacker ratios, and data distributions.
    
\end{itemize}


\vspace{-4mm}
\section{Preliminaries}
\label{sec:Preliminaries}
\vspace{-2mm}
\subsection{Split Federated Learning}
\label{sec:2.1}
In SFL~\cite{SplitFed}, the model is split into two parts at a specific layer, referred to as the \textbf{cut layer}. The segment from the input layer to the cut layer is denoted as the \textbf{bottom model}, while the segment spanning from the cut layer to the output layer is designated as the \textbf{top model}. Unlike conventional distributed paradigms (\eg, FL, SL), SFL involves the operation of two distinct servers: the \textbf{main server}, responsible for updating the top model, and the \textbf{fed server}, tasked with updating the bottom model. To elucidate further, a SFL system with $K$ clients iteratively executes the ensuing steps until convergence of the global model is achieved:

\vspace{-2mm}
\begin{itemize}
	\item \textbf{Step I:} In the $t$-th iteration, the fed server broadcasts a bottom model $\boldsymbol{w_{t}^{B}}$ to the clients.

\vspace{-2mm}
	\item \textbf{Step II:} Upon receiving $\boldsymbol{w_{t}^{B}}$, each individual client $k$ feeds the local data $D_{k}$ into $\boldsymbol{w_{t}^{B}}$, yielding \textbf{smashed data} $\boldsymbol{A_{t,k}}$. Subsequently, $\boldsymbol{A_{t,k}}$ and the corresponding labels $\boldsymbol{Y_{k}}$ are uploaded to the main server.

\vspace{-2mm}
	\item \textbf{Step III:} Based on $\boldsymbol{A_{t,k}}$, $\boldsymbol{Y_{k}}$, and the top model $\boldsymbol{w_{t}^{T}}$, the main server computes $\nabla \ell_k\left(\boldsymbol{A_{t,k}}; \boldsymbol{w_{t}^{T}},\boldsymbol{Y_{k}}\right)$ (\ie, the gradient of $\boldsymbol{A_{t,k}}$) and $\nabla \ell_k\left(\boldsymbol{w_{t}^{T}}; \boldsymbol{A_{t,k}},\boldsymbol{Y_{k}}\right)$ (\ie, the gradient of $\boldsymbol{w_{t}^{T}}$). The former is sent to client $k$ for updating the bottom model, while the latter is utilized to aggregate and update the global top model:
 \begin{equation}
	\boldsymbol{w_{t+1}^{T}}\gets \boldsymbol{w_{t}^{T}}-\eta \cdot AGR(\nabla \ell_k\left(\boldsymbol{w_{t}^{T}}; \boldsymbol{A_{t,k}},\boldsymbol{Y_{k}}\right), k\in [K])
 \end{equation}
 where AGR is an aggregation algorithm (\eg, FedAvg~\cite{FedAvg}), $\eta$ is the learning rate.

\vspace{-2mm}
    \item \textbf{Step IV:} Each client $k$ updates its local bottom model $\boldsymbol{w_{t,k}^{B}}$ with  $\nabla \ell_k\left(\boldsymbol{A_{t,k}}; \boldsymbol{w_{t}^{T}},\boldsymbol{Y_{k}}\right)$ and subsequently uploads the updated $\boldsymbol{w_{t,k}^{B}}$ to the fed server.

\vspace{-2mm}
    \item \textbf{Step V:} The fed server utilizes the aggregation algorithm AGR to amalgamate all local bottom models, thereby updating $\boldsymbol{w_{t}^{B}}$  for the next iteration:
  \begin{equation}
	\boldsymbol{w_{t+1}^{B}}\gets AGR(\boldsymbol{w_{t,k}^{B}}, k\in [K])
 \end{equation}
\end{itemize}

\vspace{-2mm}
\subsection{Threat Model}
We assume that SFL comprises a total of $K$ clients, with $M$ clients designated as malicious. To align with existing research~\cite{SFLAGRT,SFLDataPoisoning}, we adopt a default attacker proportion of $20\%$ (\ie, $\frac{M}{K}=0.2$). The attackers' objective is to minimize the accuracy of the global model by uploading poisoning parameters (\eg, smashed data, labels, and local bottom models) to the servers. To ensure the practicality of the attack, we make minimal assumptions about the attackers' knowledge. Specifically, our attack is AGR-agnostic, meaning the attackers are unaware of the aggregation algorithm deployed by the servers. We also impose significant constraints on the attackers' capabilities. In particular, they can only manipulate their own training processes, training data, and uploaded parameters. They do not possess control over the servers or the training activities of other benign clients.

\vspace{-4mm}
\section{MISA}
Existing research~\cite{SFLAGRT,SFLDataPoisoning,SFLDataPoisoning2} has demonstrated that SFL exhibits strong resilience against poisoning attacks. It is argued that the partitioning of the model significantly constrains the attackers' ability to poison, as they are limited to targeting the bottom model only. However, all existing research has overlooked the possibility of poisoning the top model through malicious smashed data. Building upon this insight,  we present a novel poisoning attack known as MISA. It poisons both the top and bottom models, causing a \textbf{\underline{misa}}lignment in the global model, ultimately leading to a drastic accuracy collapse.

\vspace{-2mm}
\subsection{Attacking Top Model}
Given that the top model is controlled exclusively by the main server and remains inaccessible to any client sharing, attackers cannot directly poison the top model. Fortunately, the updating of top model parameters relies upon the smashed data and corresponding labels uploaded by clients (please refer to \textbf{Step III} in Section~\ref{sec:2.1}). Thus, we have the opportunity to meticulously craft malicious smashed data, so as to introduce bias into the top model.

Formally, for a smashed data point $x_i$ with label $y$, we poison it as follows:

\vspace{-4mm}
\begin{equation}
\label{formula:3}
	\hat{x}_i = \lambda x_i +(1-\lambda)\Bar{x}-1.5\delta
 \end{equation}
where $\Bar{x}$ is the mean of all smashed data labeled y held by attackers, $\delta$ stands for the standard deviation of all such data, $1.5$ is a commonly used standard deviation perturbation magnitude~\cite{ALittleIsEnough}, which is sufficiently large to achieve the perturbation effect and, at the same time, small enough to evade detection by defense systems. We consistently set $\lambda$, a hyperparameter, to 0.05 in this paper.

The rationale behind formula~(\ref{formula:3}) is to identify relatively malicious values within a reasonable range using the mean and variance (smashed data can be approximated as a normal distribution). The role of $\lambda$ is to ensure that each maliciously constructed smashed data point is distinct, as failing to do so would render it easily detectable.

\vspace{-2mm}
\subsection{Attacking Bottom Model}
As illustrated in \textbf{Step V} in Section~\ref{sec:2.1}, the global bottom model is obtained from the local bottom models uploaded by all clients, allowing attackers to directly poison their own bottom models.

Inspired by \cite{AGRT}, we formalize the poisoning of the bottom model as a Min-Sum problem, where the sum of squared distances of the malicious bottom model from all the benign models is upper-bounded by the sum of squared distances of any benign bottom model from the other benign bottom models. In this way, the crafted malicious bottom model appears less like an outlier than some benign bottom models at the edge. Formally, the objective can be formalized as follows: 

\begin{equation}
\begin{gathered}
\underset{\gamma, \nabla^p}{\operatorname{argmax}} \sum_{i \in bng}\left\|\boldsymbol{w_{t,m}^{B}}-\boldsymbol{w_{t,i}^{B}}\right\|_2^2 \leq \max _{i \in bng} \sum_{j \in bng}\left\|\boldsymbol{w_{t,i}^{B}}-\boldsymbol{w_{t,j}^{B}}\right\|_2^2 \\
\boldsymbol{w_{t,m}^{B}}=f_{\text {avg }}\left(\boldsymbol{w_{t,i}^{B}},i\in bng\right)+\gamma \nabla^p
\end{gathered}
\end{equation}
where $bng$ is a set of the indices of all benign clients, $f_{avg}$ represents the FedAvg aggregation, while $\nabla^p$ and $\gamma$ correspond to the direction and magnitude of perturbation, respectively.

\noindent \textbf{Obtaining the perturbation direction. }Existing research~\cite{AGRT,LocalModelPoisoning} typically fixes the perturbation direction when optimizing this objective, focusing solely on exploring the perturbation magnitude. However, in real-world scenarios, variations in datasets, data distributions, attacker ratios, aggregation algorithms, and other factors may lead to changes in the optimal perturbation direction. 

In light of this, we propose an adaptive algorithm to search for the perturbation direction. This approach dynamically selects the optimal perturbation direction for different scenarios, thus maximizing the attack performance. Specifically, we model the choice of perturbation direction as a \textit{Multi-Armed Bandit} (MAB) problem~\cite{MABRFL,FDP} and employ the Thompson Sampling~\cite{thompsonsampling} algorithm to select the most promising perturbation direction from the three classical directions~\cite{AGRT}: inverse unit vector, inverse standard deviation, and inverse sign. Note that we configure our reward function in Thompson Sampling to measure the deviation of the global model's direction (using cosine similarity).

\noindent \textbf{Obtaining the perturbation magnitude. }Existing study~\cite{AGRT} employs oscillation optimization to search for perturbation magnitude. However, this approach explores a very narrow range ($0.5\gamma_{init} \sim 1.5\gamma_{init}$), making it challenging to set an appropriate hyperparameter $\gamma_{init}$ that can potentially find the optimal perturbation magnitude in each round. This limitation significantly compromises the effectiveness of existing attacks.

To address this, we introduce SnL (search and locate). Specifically, we start with a coarse-grained search using a binary search to locate an approximate $\gamma_{init}$. Then, we utilize the oscillation optimization method with this value as the initial point for fine-grained localization.

\begin{figure*}[t]
	\centering
	\subfigure[Impact of split positions]{\label{fig:sp}
		\includegraphics[width=0.6\columnwidth]{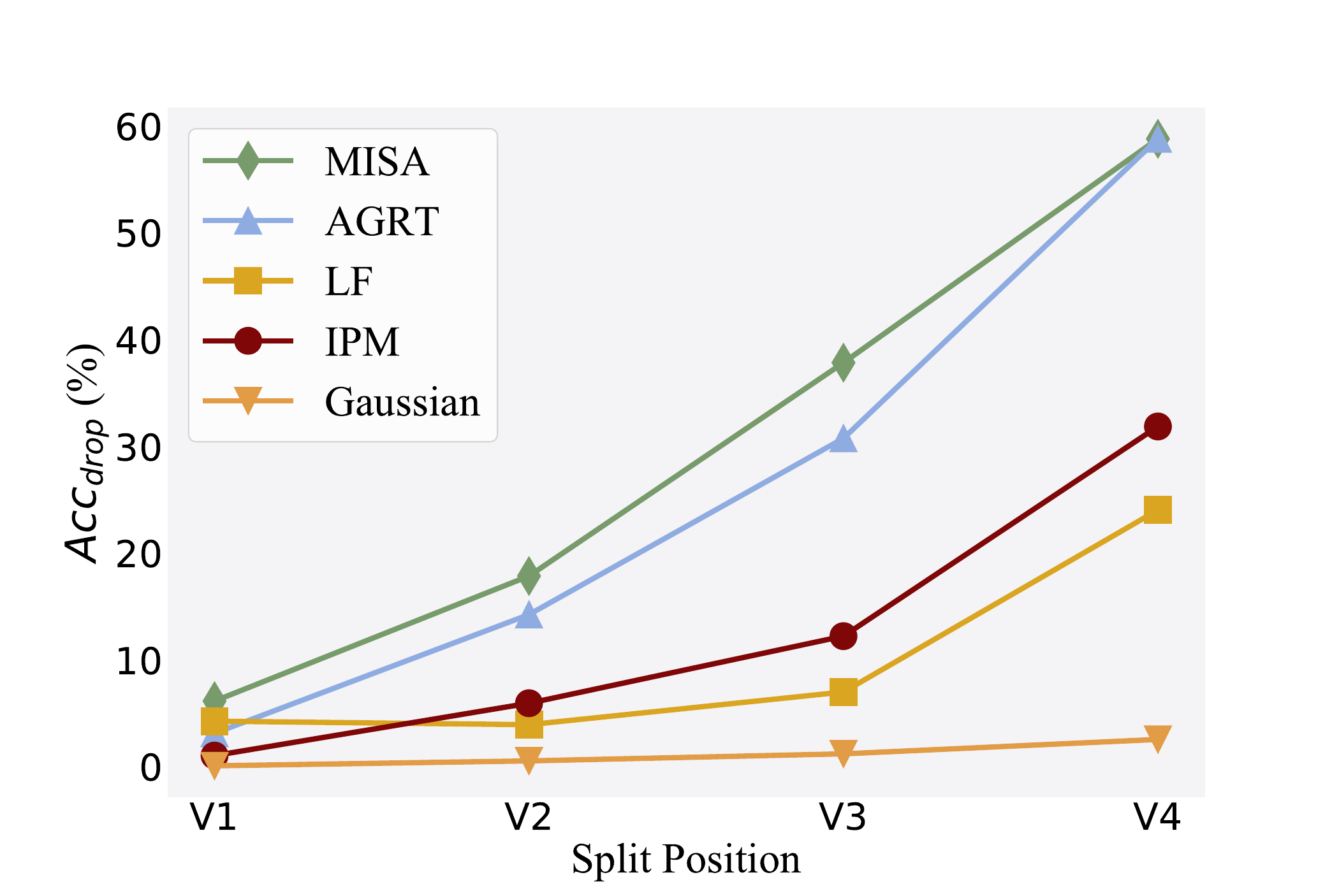}}\hspace{8mm}
	\subfigure[Impact of attacker ratios]{\label{fig:ar}
		\includegraphics[width=0.6\columnwidth]{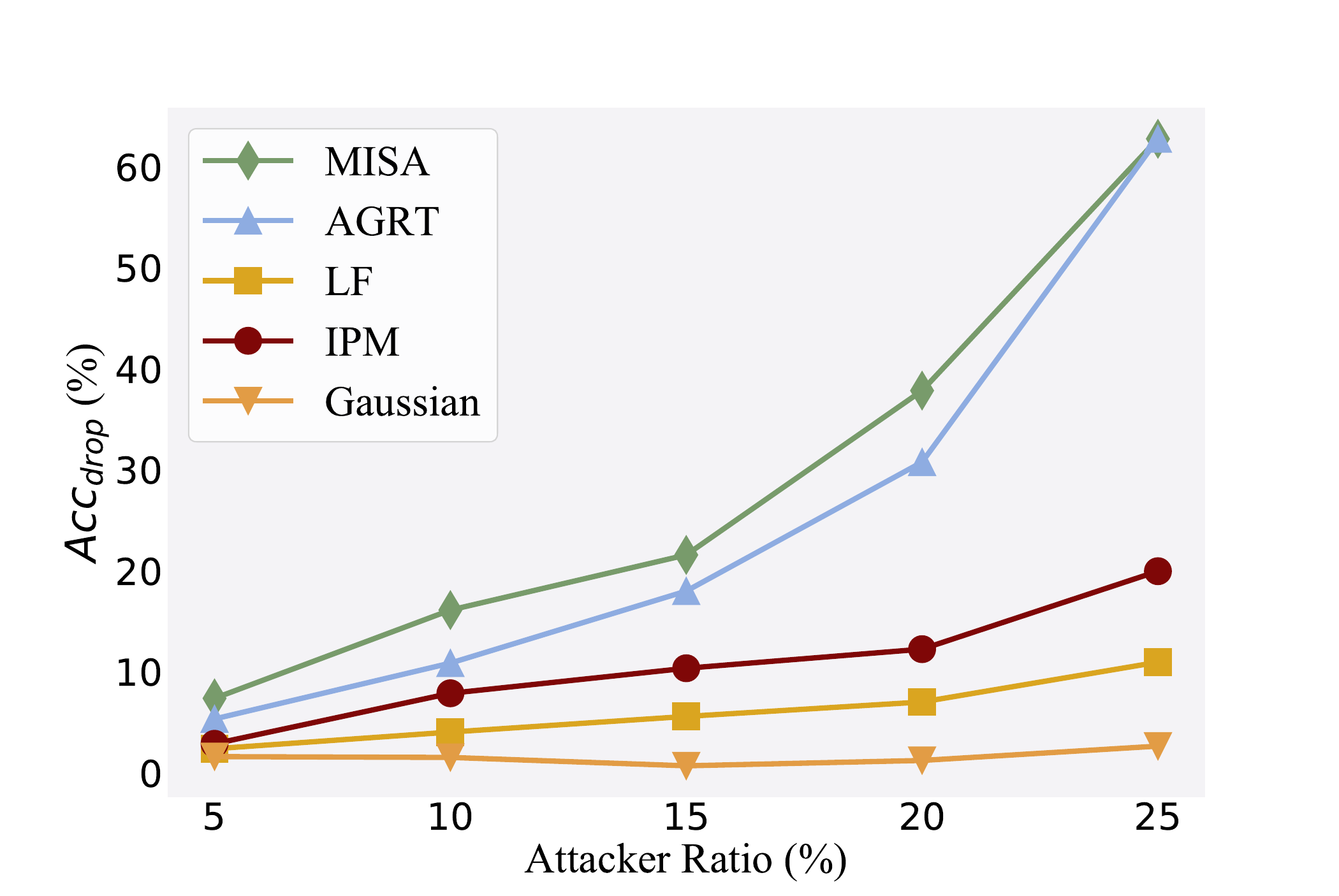}}\hspace{8mm}
	\subfigure[Impact of data distributions]{\label{fig:dd}
		\includegraphics[width=0.6\columnwidth]{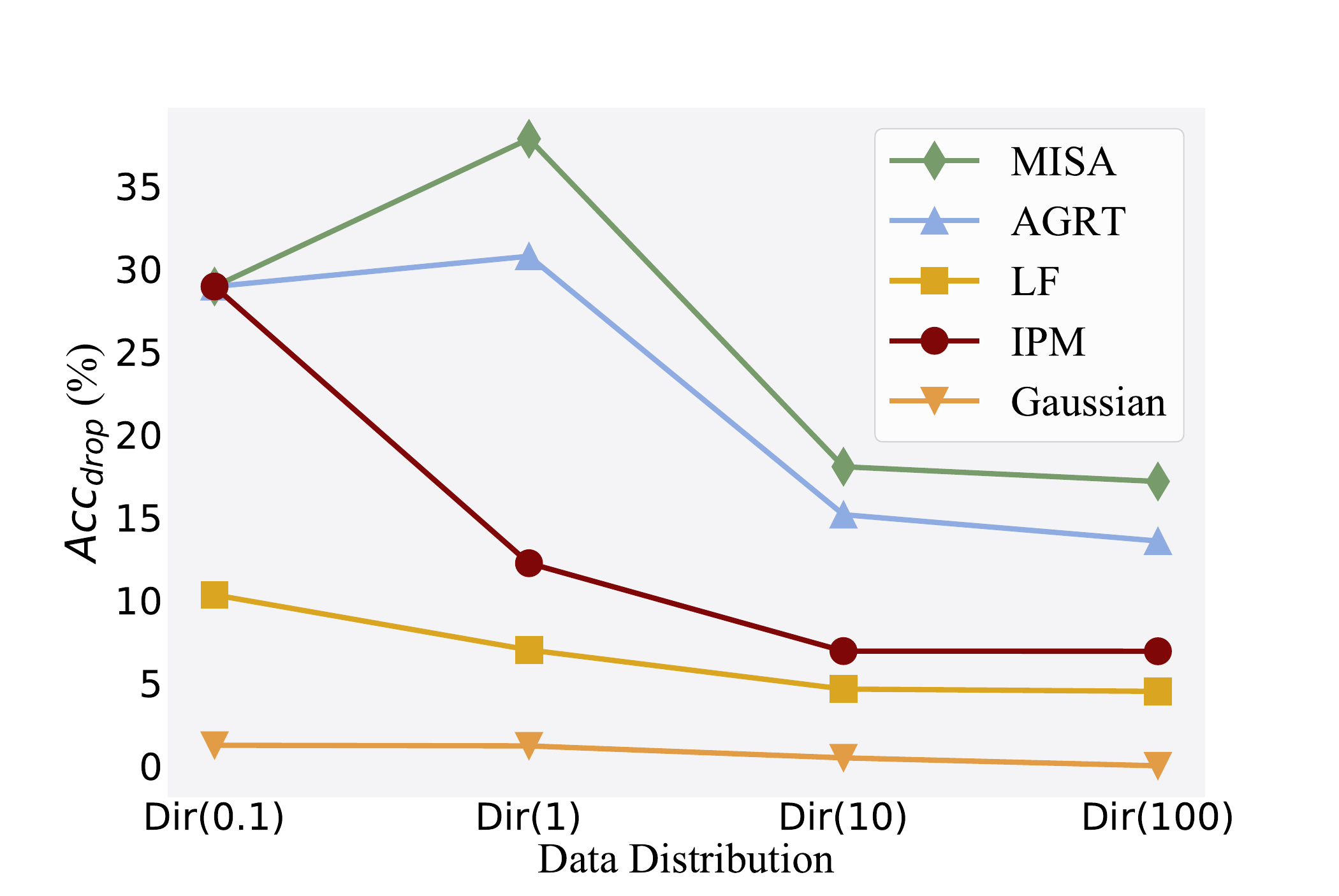}}
	\vspace{-2mm}
	\caption{Impact of some important factors on accuracy drop}
	\vspace{-5mm}
	\label{fig:impactoffactors}
\end{figure*}

\vspace{-4mm}
\section{Experiments}
\subsection{Experimental Setup}
We conduct experiments on CIFAR-10 and employ Dirichlet sampling (with $\alpha=1$) to simulate the data distribution for each client. We consider a VGG-like model (consisting of 5 convolutional layers followed by a global average pooling layer and a fully connected layer) and split it at the third convolutional layer. We set the number of clients $K=100$, of attackers $M=20$. The number of iterations per experiment is set to 200. In addition to AGRT~\cite{SFLAGRT} and LF (\textit{label flipping})~\cite{SFLDataPoisoning}, the only two poisoning attacks in SFL domain, we also consider two strong attacks from FL domain: Gaussian~\cite{FedInv} and IPM (\textit{inner product manipulation})~\cite{IPM}. For defenses, we evaluate three effective strategies: Krum~\cite{Krum}, TrMean~\cite{TrimmedMean}, and Median~\cite{TrimmedMean}. We designate Median as the default defense due to its good performance. We quantify the attack impact with accuracy drop ($Acc_{drop}$), which represents the difference in accuracy without and with the attack (denoted as $Acc$, and $Acc_{attack}$ respectively).
\subsection{Experimental Results}
\begin{table}[!t]
\renewcommand\arraystretch{1}
\centering
\caption{Comparison with SOTA attacks}
\resizebox{0.95\linewidth}{!}{
\begin{tabular}{|c|c|ccccc|}
\hline
\multirow{2}{*}{AGR} & \multirow{2}{*}{Acc (\%)} & \multicolumn{5}{c|}{$Acc_{drop}$ (\%)} \\ \cline{3-7} 
 &  & \multicolumn{1}{c|}{AGRT} & \multicolumn{1}{c|}{LF} & \multicolumn{1}{c|}{Gaussian} & \multicolumn{1}{c|}{IPM} & MISA \\ \hline
Krum & 52.13 & \multicolumn{1}{c|}{\textbf{42.13}} & \multicolumn{1}{c|}{6.87} & \multicolumn{1}{c|}{5.31} & \multicolumn{1}{c|}{4.98} & 15.19 \\ \hline
TrMean & 73.78 & \multicolumn{1}{c|}{27.77} & \multicolumn{1}{c|}{7.06} & \multicolumn{1}{c|}{1.27} & \multicolumn{1}{c|}{16.27} & \textbf{32.97} \\ \hline
Median & 72.79 & \multicolumn{1}{c|}{30.78} & \multicolumn{1}{c|}{7.03} & \multicolumn{1}{c|}{1.24} & \multicolumn{1}{c|}{12.26} & \textbf{37.86} \\ \hline
\end{tabular}
}
\label{table:Comparison with SOTA attacks}
\end{table}

\noindent \textbf{Comparison with SOTA attacks. }As depicted in Tab.~\ref{table:Comparison with SOTA attacks}, under TrMean and Median, MISA demonstrates superior performance, exhibiting significantly higher $Acc_{drop}$ values than its competitors. In the case of Krum, while its $Acc_{drop}$ may not surpass AGRT, it's important to note that AGRT requires prior knowledge of the server's deployed defense mechanisms, an assumption that is often unrealistic in real-world scenarios. Compared to the other AGR-agnostic attacks, including LF, Gaussian, and IPM, MISA outperforms them by a substantial margin.

\noindent \textbf{Impact of split positions. }We split the model at the $i$-th convolutional layer, where $i \in \left\{1, 2, 3, 4\right\}$, and denote the split position as $V_i$. As demonstrated in Fig.~\ref{fig:sp}, with the deepening of the split position, the attack effectiveness of all attacks is enhanced to varying degrees. Nevertheless, it is noteworthy that MISA consistently outperforms the others.

\noindent \textbf{Impact of attacker ratios. }Fig.~\ref{fig:ar} showcases the  $Acc_{drop}$ comparison for various attacks as the attacker ratio varies from $5\%$ to $25\%$. It is observed that LF, Gaussian, and IPM exhibit relatively stable $Acc_{drop}$ changes as the attacker ratio increases. In contrast, both AGRT and MISA demonstrate substantial increases in $Acc_{drop}$. On average, however, MISA achieves an $Acc_{drop}$ approximately $5\%$ higher than AGRT.

\noindent \textbf{Impact of data distributions. }We employ Dirichlet sampling to simulate the distribution of the data, denoted as $Dir(\alpha)$, where a larger value of $\alpha$ indicates a lower degree of data heterogeneity. Fig.~\ref{fig:dd} considers four different data distributions, and it can be observed that as the degree of data heterogeneity decreases, the $Acc_{drop}$ for all attacks reduces, indicating increased difficulty in attacking. However, MISA consistently achieves the highest $Acc_{drop}$, even under the extremely homogeneity setting (\eg, $Dir(100)$), where MISA can still cause a reduction of approximately $20\%$ in accuracy. This confirms the superiority of MISA.

\noindent \textbf{Ablation study. }In Tab.~\ref{table:Ablation study}, we consider three combinations: attacking both models simultaneously (\ie, MISA), attacking solely the top model, and attacking solely the bottom model. We are surprised to find that MISA usually yields a notably higher $Acc_{drop}$ compared to the additive $Acc_{drop}$ from separately attacking the top model and bottom model (with increases of $41.70\%$ and $34.16\%$ in $Acc_{drop}$ for Krum and Median, respectively). This suggests that model misalignment can indeed lead to a collapse in accuracy.
\begin{table}[!t]
\renewcommand\arraystretch{1}
\centering
\caption{Ablation study}
\resizebox{0.95\linewidth}{!}{
\begin{tabular}{|c|ccc|}
\hline
\multirow{2}{*}{Combination} & \multicolumn{3}{c|}{$Acc_{drop}$   (\%)} \\ \cline{2-4} 
 & \multicolumn{1}{c|}{Krum} & \multicolumn{1}{c|}{TrMean} & Median \\ \hline
MISA & \multicolumn{1}{c|}{15.19} & \multicolumn{1}{c|}{32.97} & 37.86 \\ \hline
Attacking top model & \multicolumn{1}{c|}{5.90} & \multicolumn{1}{c|}{4.57} & 5.56 \\ \hline
Attacking bottom model & \multicolumn{1}{c|}{4.82} & \multicolumn{1}{c|}{29.96} & 22.66 \\ \hline
\end{tabular}
}
\label{table:Ablation study}
\end{table}


\vspace{-4mm}
\section{Acknowledgments} 
\vspace{-2mm}
Shengshan's work is supported in part by the National Natural Science Foundation of China (Grant No.U20A20177) and Hubei Province Key R\&D Technology Special Innovation Project under Grant No.2021BAA032. 
Minghui's work is supported in part by the National Natural Science Foundation of China (Grant No. 62202186). The work is supported by HPC Platform of Huazhong University of Science and Technology.
Shengshan Hu is the corresponding author. 


\bibliographystyle{IEEEbib}
\bibliography{strings,refs}

\begin{thebibliography}{10}

\bibitem{FedAvg}
Brendan McMahan, Eider Moore, Daniel Ramage, Seth Hampson, and
  Blaise~Ag{\"{u}}era y~Arcas,
\newblock ``Communication-efficient learning of deep networks from
  decentralized data,''
\newblock in {\em Proceedings of the 20th International Conference on
  Artificial Intelligence and Statistics ({AISTATS}'17)}, 2017, vol.~54, pp.
  1273--1282.

\bibitem{FLSurvey}
Junyu Shi, Wei Wan, Shengshan Hu, Jianrong Lu, and Leo~Yu Zhang,
\newblock ``Challenges and approaches for mitigating byzantine attacks in
  federated learning,''
\newblock in {\em Proceedings of {IEEE} International Conference on Trust,
  Security and Privacy in Computing and Communications (TrustCom'22)}, 2022,
  pp. 139--146.

\bibitem{SL}
Praneeth Vepakomma, Otkrist Gupta, Tristan Swedish, and Ramesh Raskar,
\newblock ``Split learning for health: Distributed deep learning without
  sharing raw patient data,''
\newblock {\em CoRR}, vol. abs/1812.00564, 2018.

\bibitem{FLDPGAN}
Longling Zhang, Bochen Shen, Ahmed Barnawi, Shan Xi, Neeraj Kumar, and Yi~Wu,
\newblock ``Feddpgan: Federated differentially private generative adversarial
  networks framework for the detection of {COVID-19} pneumonia,''
\newblock {\em Inf. Syst. Frontiers}, vol. 23, no. 6, pp. 1403--1415, 2021.

\bibitem{SplitFed}
Chandra Thapa, Mahawaga Arachchige~Pathum Chamikara, Seyit Camtepe, and Lichao
  Sun,
\newblock ``Splitfed: When federated learning meets split learning,''
\newblock in {\em Proceedings of the 36th {AAAI} Conference on Artificial
  Intelligence ({AAAI}'22)}, 2022, pp. 848--8493.

\bibitem{SFLAGRT}
Momin~Ahmad Khan, Virat Shejwalkar, Amir Houmansadr, and Fatima~M. Anwar,
\newblock ``Security analysis of splitfed learning,''
\newblock in {\em Proceedings of the 20th {ACM} Conference on Embedded
  Networked Sensor Systems (SenSys'22)}, 2022, pp. 987--993.

\bibitem{SFLDataPoisoning}
Saurabh Gajbhiye, Priyanka Singh, and Shaifu Gupta,
\newblock ``Data poisoning attack by label flipping on splitfed learning,''
\newblock in {\em Recent Trends in Image Processing and Pattern Recognition -
  5th International Conference ({RTIP2R}'22)}, 2022, vol. 1704, pp. 391--405.

\bibitem{SFLDataPoisoning2}
Aysha Thahsin~Zahir Ismail and Raj~Mani Shukla,
\newblock ``Analyzing the vulnerabilities in splitfed learning: Assessing the
  robustness against data poisoning attacks,''
\newblock {\em CoRR}, vol. abs/2307.03197, 2023.

\bibitem{thompsonsampling}
Shipra Agrawal and Navin Goyal,
\newblock ``Analysis of thompson sampling for the multi-armed bandit problem,''
\newblock in {\em Proceedings of the 25th Annual Conference on Learning Theory
  (COLT'12)}, 2012, pp. 39.1--39.26.

\bibitem{ALittleIsEnough}
Gilad Baruch, Moran Baruch, and Yoav Goldberg,
\newblock ``A little is enough: Circumventing defenses for distributed
  learning,''
\newblock in {\em Processings of the 33rd Annual Conference on Neural
  Information Processing Systems (NeurIPS'19)}, 2019, pp. 8632--8642.

\bibitem{AGRT}
Virat Shejwalkar and Amir Houmansadr,
\newblock ``Manipulating the byzantine: Optimizing model poisoning attacks and
  defenses for federated learning,''
\newblock in {\em Proceedings of the 28th Annual Network and Distributed System
  Security Symposium ({NDSS}'21)}, 2021.

\bibitem{LocalModelPoisoning}
Minghong Fang, Xiaoyu Cao, Jinyuan Jia, and Neil~Zhenqiang Gong,
\newblock ``Local model poisoning attacks to byzantine-robust federated
  learning,''
\newblock in {\em Proceedings of the 29th {USENIX} Security Symposium ({USENIX}
  Security'20)}, 2020, pp. 1605--1622.

\bibitem{MABRFL}
Wei Wan, Shengshan Hu, Jianrong Lu, Leo~Yu Zhang, Hai Jin, and Yuanyuan He,
\newblock ``Shielding federated learning: Robust aggregation with adaptive
  client selection,''
\newblock in {\em Proceedings of the 31st International Joint Conference on
  Artificial Intelligence ({IJCAI}'22)}, 2022, pp. 753--760.

\bibitem{FDP}
Wei Wan, Shengshan Hu, Minghui Li, Jianrong Lu, Longling Zhang, Leo~Yu Zhang,
  and Hai Jin,
\newblock ``A four-pronged defense against byzantine attacks in federated
  learning,''
\newblock in {\em Proceedings of the 31st {ACM} International Conference on
  Multimedia, ({MM}'23)}, 2023, pp. 7394--7402.

\bibitem{FedInv}
Bo~Zhao, Peng Sun, Tao Wang, and Keyu Jiang,
\newblock ``Fedinv: Byzantine-robust federated learning by inversing local
  model updates,''
\newblock in {\em Proceedings of the 36th {AAAI} Conference on Artificial
  Intelligence ({AAAI}'22)}, 2022, pp. 9171--9179.

\bibitem{IPM}
Cong Xie, Oluwasanmi Koyejo, and Indranil Gupta,
\newblock ``Fall of empires: Breaking byzantine-tolerant sgd by inner product
  manipulation,''
\newblock in {\em Proceedings of the 36th International Conference on
  Uncertainty in Artificial Intelligence (UAI'20)}, 2020, pp. 261--270.

\bibitem{Krum}
Peva Blanchard, El~Mahdi~El Mhamdi, Rachid Guerraoui, and Julien Stainer,
\newblock ``Machine learning with adversaries: Byzantine tolerant gradient
  descent,''
\newblock in {\em Proceedings of the 31st Annual Conference on Neural
  Information Processing Systems (NeurIPS'17)}, 2017, pp. 119--129.

\bibitem{TrimmedMean}
Dong Yin, Yudong Chen, Kannan Ramchandran, and Peter~L. Bartlett,
\newblock ``Byzantine-robust distributed learning: Towards optimal statistical
  rates,''
\newblock in {\em Proceedings of the 35th International Conference on Machine
  Learning ({ICML}'18)}, 2018, vol.~80, pp. 5636--5645.

\end{thebibliography}
\end{document}